\documentclass{article}
\usepackage{graphicx}
\usepackage{amsmath}
\usepackage{booktabs}
\usepackage{multirow}
\usepackage{adjustbox}
\usepackage{subcaption}
\usepackage{float}
\usepackage[round]{natbib}
\usepackage[colorlinks, citecolor=blue]{hyperref}
\usepackage[a4paper,
            left=3cm,
            right=3cm,
            top=3cm,
            bottom=3cm]{geometry}
\title{Toward Better Generalisation in Uncertainty Estimators: Leveraging Data-Agnostic Features}

\author{
    \textbf{Thuy An Ha} \\
    Swinburne University of Technology \\
    104697626@student.swin.edu.au
    \and
    \textbf{Bao Quoc Vo} \\
    Swinburne University of Technology \\
    bvo@swin.edu.au
}

\date{}

\begin{document}

\maketitle

\centerline{\small \textbf{Abstract}}

\vspace{1em}

Large Language Models (LLMs) often generate responses that are factually incorrect yet expressed with high confidence, which can pose serious risks for end users. To address this, it is essential for LLMs not only to produce answers but also to provide accurate estimates of their correctness. Uncertainty quantification methods have been introduced to assess the quality of LLM outputs, with factual accuracy being a key aspect of that quality. Among these methods, those that leverage hidden states to train probes have shown particular promise, as these internal representations encode information relevant to the factuality of responses, making this approach the focus of this paper. However, the probe trained on the hidden states of one dataset often struggles to generalise to another dataset of a different task/domain. To address this limitation, we explore combining data-agnostic features with hidden‑state features and assess whether this hybrid feature set enhances out‑of‑domain performance. We further examine whether selecting only the most informative hidden‑state features, thereby discarding task‑specific noise, enables the data‑agnostic features to contribute more effectively. The experiment results indicate that although introducing data‑agnostic features generally enhances generalisation performance in most cases, in certain scenarios their inclusion degrades performance. A similar pattern emerges when retaining only the most important hidden‑state features - adding data‑agnostic features does not consistently further enhance performance compared to using the full set of hidden-state features. A closer analysis reveals that, in some specific cases, the trained probe underweights the data-agnostic features relative to the hidden-state features, which we believe is the main reason why the results are inconclusive. Our code and all complementary documents can be found at \href{https://github.com/HaThuyAn/ResearchProject}{this GitHub repository}.

\section{Introduction}
In recent years, Large Language Models (LLMs) have demonstrated remarkable capabilities across a variety of tasks \citep{Touvron2023llama2, Achiam2023gpt4, Chowdhery2023palm}. However, LLMs tend to produce non-factual responses with convincing fluency, restricting their adoption in real‑world applications, especially in safety‑critical domains such as healthcare \citep{Salvagno2023hallucinations}; \citep{Huang2023survey}. In the absence of explicit cautions, users may blindly accept model outputs, potentially leading to serious consequences. Therefore, robust uncertainty estimation methods are essential to quantify response reliability, providing a reliable clue for when the output should or should not be trusted. \\

Recent research suggests that the internal states of the LLMs carry information that is highly correlated with truthfulness, allowing uncertainty to be estimated from these representations, where higher uncertainty generally corresponds to a lower likelihood of factual correctness \citep{Azaria2023internalstate, Beigi2024internalinspector, Orgad2024intrinsic}. Studies have investigated various methods that utilise the hidden states of the LLMs to predict uncertainty. A widely used approach involves training a probe directly on the hidden-state activations produced during text generation. While this method has demonstrated strong performance compared to techniques such as verbalised confidence, token or sequence probability, and self-consistency \citep{Mahaut2024factual, Liu2024uncertainty}, it suffers from a major drawback of being unable to generalise well across different tasks and datasets \citep{Azaria2023internalstate, Beigi2024internalinspector, Ulmer2024calibrating, Liu2024hyperplane, Orgad2024intrinsic}, implying that hidden states capture task‑specific rather than universal information \citep{Orgad2024intrinsic}. Two lines of work have approached this issue by proposing different techniques. Specifically, \cite{Liu2024hyperplane} proposed to train the probe on a set of multiple datasets covering different tasks and domains, yielding a decent generalisation performance. They suggested that when increasing the diversity of the training data, the truthfulness hyperplane can be found and utilised to reliably predict uncertainty even in cross-task setting. However, this method requires collecting additional training data from different tasks and domains, which is resource intensive. On the other hand, \cite{Zhang2024promptguided} argued that truthfulness information is entangled with domain‑specific content as hidden states are optimised for text generation. Therefore, they have suggested a technique of training the probe on only one dataset, which can still achieve high generalisation performance, by modifying the prompt to include the statement and a question asking whether the statement is correct. However, this work is limited to only True/False statements in the scope of different topics without examining datasets of different tasks and domains. In this paper, we propose a method that does not require heavy multi-dataset training and also goes beyond the True/False format, enabling robust, cross-domain uncertainty estimation without requiring extensive training data or being limited to binary statements. Inspired by the concept of data-agnostic features which are proposed to be effective in improving generalisation performance \citep{He2024factoscope}\footnote{Unlike the work from \cite{He2024factoscope} which only considers generalisation for different topics such as art, history, sports, etc. Our work is different in the sense that we examine the effectiveness of the data-agnostic features across datasets spanning different tasks/domains, each requiring distinct reasoning skills from the LLMs.} as they are task-independent, we aim to answer two questions in this paper:

\begin{enumerate}
    \item Can supplementing hidden‑state features with data‑agnostic features improve the probe’s generalisation to unseen tasks or datasets?
    \item Does pruning to the top‑ranked hidden‑state features amplify this generalisation?
\end{enumerate}
Driven by these two research questions, we adopt the work from \cite{Liu2024uncertainty} as the backbone of our research, training probes on hidden states drawn from three different tasks – Factual Question Answering, Reading Comprehension, and Commonsense Reasoning, and then incorporating the data‑agnostic features to address the first research question. While \cite{Liu2024uncertainty} focused on demonstrating that grey-box features such as entropy and probability are insufficient for accurate uncertainty estimation, and that incorporating hidden state features improves performance, our research shifts the focus toward the generalisation problem. Specifically, we leverage grey-box features from both \cite{Liu2024uncertainty} and \cite{Manakul2023selfcheckgpt} as data-agnostic inputs, alongside hidden state features, and employ a similar random forest regressor as the uncertainty estimator to examine whether these data-agnostic features can help enhance generalisation performance across tasks.  The experiment results show that data-agnostic features do help improve generalisation performance in most of the cases, however, there are some situations that they fail to do so. Going further, based on the hypothesis of hidden states of the LLMs are task-specific \citep{Orgad2024intrinsic}, we investigate whether selecting only the top-ranked hidden-state features can amplify the generalisation performance as the number of task-specific features are reduced. The experiment results show that reducing and retaining only the most important data-agnostic features does help amplify the generalisation performance, however, this technique fails in certain cases. A close examination of feature‐importance scores shows that in successful cases, data-agnostic features consistently outrank the selected hidden-state features, amplifying their impact while still leveraging hidden‐state information, whereas failures occur when data-agnostic features are intermixed with lower-ranked hidden-state features and thus receive insufficient weight. \\

In summary, the contributions of this paper are as follows: (1) We train the probes on the hidden states from different datasets covering three main tasks of Factual QA, Reading Comprehension, and Commonsense Reasoning, supplementing with the data-agnostic features, aiming to investigate whether the data-agnostic features help enhance generalisation performance. (2) We employ the feature selection method to retain only the top-ranked hidden-state features to examine whether it helps amplify the generalisation performance and also identify conditions under which this strategy succeeds or fails.

\section{{Methodology}}
\subsection{{Research Question 1}}
For research question 1, we start by obtaining the generated answers from the LLMs when prompting them with questions from different datasets.  For each dataset, we construct
\[
D = \bigl\{(\theta_i,\,c_i)\bigr\}_{i=1}^n
\]
where \(n\) is the number of samples in the original dataset, \(\theta_i\) is the 1-layer hidden state produced by feeding the prompt–answer pair \((x_i, y_i')\) back into the LLM, and \(c_i\in\{0,1\}\) indicates whether the generated answer is true (\(1\)) or false (\(0\)). \\ 

Each \(D\) is split into \(D_{\mathrm{train}}\) and \(D_{\mathrm{test}}\).  We train one probe per dataset on \(D_{\mathrm{train}}\) and evaluate it on every other dataset’s \(D_{\mathrm{test}}\), establishing the baseline cross‐dataset performance using hidden states only. \\

We then extend every sample with the set of data-agnostic features \(\{\Delta_i\}\), resulting in
\[
D' = \bigl\{(\theta_i,\,\{\Delta_i\},\,c_i)\bigr\}_{i=1}^n.
\] \\
Similarly, upon splitting each \(D'\) into \(D'_{\mathrm{train}}\) and \(D'_{\mathrm{test}}\), we retrain the probes on \(D'_{\mathrm{train}}\) and re-run the same cross-dataset evaluations on all \(D'_{\mathrm{test}}\), thereby obtaining the cross-dataset performance when supplementing the hidden states with the data-agnostic features. \\

The resulting performance allows us to quantify how much, if at all, adding \(\{\Delta_i\}\) improves the probe’s ability to generalise to unseen tasks and datasets, relative to the hidden-state only baseline.

\subsection{{Research Question 2}}
For research question 2, we employ feature selection to retain only the top-$k$ important hidden‐state features from the original 1‐layer hidden‐state features.  Similarly, we have
\[
D_s = \{(\theta_{si},\,c_i)\}_{i=1}^n
\quad\text{and}\quad
D'_s = \{(\theta_{si},\,\{\Delta_i\},\,c_i)\}_{i=1}^n
\]
where $\theta_{si}$ is the selected hidden‐state features. \\

In addition, we also consider utilising 5‐layer hidden states, thereby having
\[
D_{ml} = \{(\theta_{mli},\,c_i)\}_{i=1}^n
\quad\text{and}\quad
D'_{ml} = \{(\theta_{mli},\,\{\Delta_i\},\,c_i)\}_{i=1}^n
\]
(“ml” is short for “multi‐layer”). \\

In general, along with the original 1‐layer baseline, we now have three hidden‐state configurations - 1‐layer, selected, and 5‐layer - each evaluated both with and without the data‐agnostic features.  Theoretically, the 5‐layer configuration is expected to yield the least generalisation improvement when incorporating data‐agnostic features among the three, and the selected configuration is expected to yield the largest improvement. \\

To examine if this assumption holds, we calculate the gap between using only the hidden‐state features and using the hidden‐state + data‐agnostic features.  The configuration with the largest gap (or $\Delta\mathrm{Perf}$) is deemed to have the greatest generalisation improvement.  Formally, for each configuration, we compute:
\[
\Delta\mathrm{Perf}
\;=\;
\mathrm{Perf}(\text{hidden} + \Delta)\;-\;\mathrm{Perf}(\text{hidden}).
\]

\section{{Experiment}}
\subsection{{LLMs}}
For our numerical experiments, we employ two popular open-source LLMs – Llama2-7B \citep{Touvron2023llama2} and Mistral-7B \citep{Jiang2023mistral7b}. Both compose 32 transformer layers and have 4096 hidden units per layer.

\subsection{{Tasks and Datasets }}
Three critical task categories are considered for this study – Factual QA, Commonsense Reasoning and Reading Comprehension. Each category will consider one multiple-choice question answering dataset and one short-form question answering dataset. Specifically, for Factual QA, we employ MMLU dataset \citep{Hendrycks2020mmlu} and TriviaQA dataset \citep{Joshi2017triviaqa} - no-context version. For Commonsense Reasoning, SWAG dataset \citep{Zellers2018swag} and Winogrande dataset \citep{Sakaguchi2021winogrande} will be utilised. For Reading Comprehension, RACE dataset \citep{Lai2017race} and Squad dataset \citep{Rajpurkar2016squad} will be used. 

\subsection{Extracting Hidden States }
\textbf{Preliminary: Hidden States in Transformer-based LLMs}

Hidden states are the layer-wise vector representations that a Transformer-based LLM produces as it processes an input. First, the input tokens are converted into continuous embeddings, combining token and positional vectors, before entering the model’s fixed sequence of Transformer blocks (or Transformer layers). Within each block, multiple attention heads propagate contextual information from earlier tokens to the current position, and a feed-forward (MLP) sublayer then transforms those attended features, performing computations, reasoning, and lookups. After each block, a new hidden-state representation is produced (and that post-block state is what we use in our research, though others examine states after individual attention or MLP layers). Finally, the hidden state from the last block is unembedded and projected into the vocabulary space to predict the next token. \\

\textbf{Selecting Hidden States}

Prior studies have demonstrated that hidden states from the middle layers yield the best performance \citep{Liu2024uncertainty, Azaria2023internalstate, Duan2024hallucination}. In this study, we conduct experiments using the 15th layer (0-indexed) and utilising only the last token from answer as suggested by Liu et al. (2024). For multi-layer case, we used 5 middle layers starting from layer 13 to layer 17 (inclusive).

\subsection{Obtaining Labels}

For multiple-choice question answering, labels are obtained by simply comparing the ground truth answers given by the dataset with the generated answers by the LLMs. If the generated answer is the same as the ground truth answer, the label will be 1, otherwise, the label will be 0. \\

For short-form question answering, labels are obtained using the automatic evaluation technique by utilising the Rouge-L metric \citep{Lin2004rouge}. Specifically, the results of Rouge-L will be the labels.

\subsection{Data-agnostic Features}
\subsubsection*{Multiple‐choice question answering datasets}

For the three datasets – MMLU, RACE and SWAG, which are in the form of multiple‐choice questions and the answer has only one token from \(\{A, B, C, D\}\), follow \cite{Liu2024uncertainty}, we utilised 5 data‐agnostic features. Specifically, we will take the output logits of each token from \(\{A, B, C, D\}\) (denoted as \(z_A, z_B, z_C\) and \(z_D\)) and apply the softmax function to obtain the probability for each, which is as follows:
\[
p(y \mid x)
=
\frac{e^{z_y}}
{\sum_{y'\in\{A,B,C,D\}} e^{z_{y'}}}
\;,\quad
y\in\{A,B,C,D\}.
\]
As a result, we will have the first 4 data‐agnostic features \(\{p_A, p_B, p_C, p_D\}\). We will then calculate the entropy of this distribution as follows:
\[
H = - \sum_{y'\in\{A,B,C,D\}} p_{y'} \,\log\bigl(p_{y'}\bigr).
\]
Eventually, we will have 5 data‐agnostic features – 4 probability values sorted in descending order and 1 entropy.

\subsubsection*{Short‐form question answering datasets}

For the three datasets – TriviaQA, Squad, and Winogrande, which have short‐form answers, we utilised 4 data‐agnostic features from \cite{Manakul2023selfcheckgpt}. Specifically, we will have two probability‐based metrics and two entropy‐based metrics. Let \(N\) denote the number of tokens in the response, let \(n\) denote the \(n\)th token in the response, and let \(p_n\) be the probability of the \(n\)th token. The probability‐based metrics can be calculated as follows:
\[
\mathrm{Avg}(-\log p) \;=\; -\frac{1}{N}\sum_{n}\log p_n,
\quad
\mathrm{Max}(-\log p) \;=\; \max_{n}\bigl(-\log p_n\bigr).
\]
The entropy of the output distribution is as follows, where \(W\) is the set of all possible tokens in the vocabulary, \(p_n(w)\) is the probability of token \(w\) being generated at the \(n\)th position, and \(H_n\) is the entropy at the \(n\)th position in the response:
\[
H_n = -\sum_{w\in W} p_n(w)\,\log p_n(w).
\]
Similar to the probability‐based metrics, two entropy‐based metrics can be calculated as follows:
\[
\mathrm{Avg}(H) \;=\; \frac{1}{N}\sum_{n}H_n,
\quad
\mathrm{Max}(H) \;=\; \max_{n}H_n.
\]
Eventually, we will have 4 data‐agnostic features – two probability‐based metrics and two entropy‐based metrics.

\subsection{Supervised Probe}

Various model architectures were implemented in prior studies. For example, \cite{Azaria2023internalstate} utilised a simple feedforward neural network with 3 hidden layers with ReLU activation function and a sigmoid output. Similarly, \cite{Liu2024uncertainty} also employed a simple Random Forest Regressor model. In contrast, \cite{Beigi2024internalinspector} and \cite{He2024factoscope} designed a more comprehensive model architecture, specifically, \cite{Beigi2024internalinspector} proposed a model composing a CNN as an encoder and an MLP as a decoder, \cite{He2024factoscope} proposed a model composing three CNNs for activation maps, top-k output indices and top-k output probabilities and a GRU network for final output ranks to produce the corresponding embeddings which are integrated through a linear layer to form a comprehensive mixed representation. \\

In this study, we build on the approach of \cite{Liu2024uncertainty} and utilise a simple Random Forest Regressor to highlight the limitations of the trained probe method and support us in seeking answers for the two research questions.

\subsection{Three Main Configurations for RQ2}

When no additional data-agnostic features are included, the model relies solely on 4096 features representing the hidden units from the middle layer of the last token of the answer. When data-agnostic features are incorporated, the total number of features increases to (4096 hidden unit features + m data-agnostic features).  \\

Additionally, we apply feature selection by choosing the top 300 features based on their absolute Pearson Correlation Coefficient, reducing the number of hidden units used for training from 4096 to 300. We then conduct experiments under two conditions: one where data-agnostic features are excluded (resulting in 300 features for training) and another where they are included (yielding a total of 300 + m features for training).  \\

Furthermore, we also utilise the hidden units from 5 middle layers (from layer 13 to 17), which results in a total of 20480 features. Two conditions are also investigated in this case, one where data-agnostic features are excluded (resulting in 20480 features for training) and another where they are included (yielding a total of 20480 + m data-agnostic features).  \\

m is the number of data-agnostic features depending on the type of tasks, for multiple-choice question answering, m = 5, for short-form question answering, m = 4.

\subsection{Evaluation Metrics}

\paragraph{Accuracy (Acc):} This metric measures the proportion of instances in the test set that are predicted correctly. Specifically, an output of the LLM is considered correct (is set to 1) if the estimated confidence score from the trained estimator exceeds a predefined threshold and incorrect (is set to 0) if it falls below. For short‐form question answering datasets, the labels are also set to 0 and 1 using the predefined threshold. These predicted results will then be compared to the ground truth labels \(c_i\) in the test set \(D_{\mathrm{test}} = \{(\theta_i, c_i)\}_{i=1}^n\). Accuracy is calculated as follows:
\[
\mathrm{Acc}
\;=\;
\frac{1}{\lvert D_{\mathrm{test}}\rvert}
\sum_{i=1}^{\lvert D_{\mathrm{test}}\rvert}
\mathbf{1}(\hat c_i = c_i).
\] \\
In this study, we set the threshold at 0.5 following \cite{Azaria2023internalstate} and \cite{Beigi2024internalinspector} throughout our experiments, unless stated otherwise.

\paragraph{Area Under the Receiver Operating Characteristic Curve (ROC AUC or AUROC):} This metric is used to measure the model’s ability to distinguish between correct and incorrect predictions.

\paragraph{Expected Calibration Error (ECE):} This metric is used to measure the calibration performance of the model. \\

In this study, we mainly focus on the Accuracy metric but still provide the results of ROC AUC and ECE for reference. Specifically, in most cases, the improvements in ROC AUC are consistent with those observed for Accuracy. In terms of ECE values, they fall into a very narrow range of \((0, 0.2)\), making differences between methods negligible. In addition, post‐calibration methods can be integrated to yield better ECE performance as suggested by \cite{Liu2024uncertainty}, however, it is not in the scope of our study.

\subsection{Experimental Results and Discussion}

Notations: \\
\textit{When we train the probe on one dataset (e.g., dataset A) and test it on that same dataset, we denote that process as “A”.} \\
\textit{When we train the probe on one dataset (e.g., dataset A) and test it on another (e.g., dataset B), we denote that process as “B-A”.} \\
\textit{When we train the probe on a combination of different datasets (e.g., dataset A and B) and test it on another (e.g., dataset C), we denote that process as “C-A\&B”.} \\

\textbf{Probe fails to generalise when being trained on hidden-state features only}
\begin{table}[H]
    \centering
    \caption{Poor cross-task generalisation performance using only 1-layer-hidden-state features on multiple-choice question answering datasets using LLama2-7B as the LLM.}
    \begin{tabular}{lccc}
        \toprule
        \textbf{Dataset(s)} & \textbf{Acc} & \textbf{ROC AUC} & \textbf{ECE} \\
        \midrule
        MMLU                         & 0.6580 & 0.7181 & 0.0484 \\
        RACE                         & 0.6795 & 0.7459 & 0.0805 \\
        SWAG                         & 0.7070 & 0.7763 & 0.1251 \\
        MMLU--RACE                   & 0.6095 & 0.6629 & 0.0761 \\
        RACE--MMLU                   & 0.5960 & 0.6714 & 0.0940 \\
        RACE--SWAG                   & 0.5160 & 0.5197 & 0.0144 \\
        SWAG--RACE                   & 0.5295 & 0.5339 & 0.0221 \\
        MMLU--SWAG                   & 0.5170 & 0.5473 & 0.0155 \\
        SWAG--MMLU                   & 0.5255 & 0.5472 & 0.0128 \\
        MMLU--SWAG\&RACE             & 0.6295 & 0.6815 & 0.0890 \\
        RACE--MMLU\&SWAG             & 0.5900 & 0.6839 & 0.1136 \\
        SWAG--MMLU\&RACE             & 0.5270 & 0.5476 & 0.0303 \\
        \bottomrule
    \end{tabular}
    \label{tab:metrics}
\end{table}

\begin{table}[H]
    \centering
    \caption{Poor cross-task generalisation performance using only 1-layer-hidden-state features on short-form question answering datasets using LLama2-7B as the LLM.}
    \begin{tabular}{lccc}
        \toprule
        \textbf{Dataset(s)} & \textbf{Acc} & \textbf{ROC AUC} & \textbf{ECE} \\
        \midrule
        TriviaQA                          & 0.7205 & 0.7960 & 0.1396 \\
        Squad                             & 0.7093 & 0.7978 & 0.1163 \\
        Winogrande                        & 0.6630 & 0.7772 & 0.1264 \\
        TriviaQA--Squad                   & 0.6350 & 0.6690 & 0.0822 \\
        TriviaQA--Winogrande              & 0.5135 & 0.5646 & 0.0427 \\
        Squad--TriviaQA                   & 0.5959 & 0.7328 & 0.1428 \\
        Squad--Winogrande                 & 0.5221 & 0.5705 & 0.0322 \\
        Winogrande--TriviaQA              & 0.5020 & 0.5489 & 0.0891 \\
        Winogrande--Squad                 & 0.5310 & 0.5529 & 0.0170 \\
        TriviaQA--Squad\&Winogrande       & 0.6170 & 0.6587 & 0.0750 \\
        Squad--TriviaQA\&Winogrande       & 0.6064 & 0.7352 & 0.1424 \\
        Winogrande--TriviaQA\&Squad       & 0.5320 & 0.5688 & 0.0394 \\
        \bottomrule
    \end{tabular}
    \label{tab:triviaqa_squad_winogrande}
\end{table}

\begin{table}[H]
    \centering
    \caption{Poor cross-task generalisation performance using only 1-layer-hidden-state features on multiple-choice question answering datasets using Mistral-7B as the LLM.}
    \begin{tabular}{lccc}
        \toprule
        \textbf{Dataset(s)} & \textbf{Acc} & \textbf{ROC AUC} & \textbf{ECE} \\
        \midrule
        MMLU                         & 0.6970 & 0.7622 & 0.0616 \\
        RACE                         & 0.7170 & 0.7863 & 0.1252 \\
        SWAG                         & 0.7545 & 0.8352 & 0.1194 \\
        MMLU--RACE                   & 0.6725 & 0.7349 & 0.1284 \\
        RACE--MMLU                   & 0.5870 & 0.7307 & 0.1390 \\
        RACE--SWAG                   & 0.5400 & 0.6401 & 0.0772 \\
        SWAG--RACE                   & 0.5405 & 0.6494 & 0.0998 \\
        MMLU--SWAG                   & 0.5700 & 0.6397 & 0.0754 \\
        SWAG--MMLU                   & 0.6145 & 0.6635 & 0.0877 \\
        MMLU--SWAG\&RACE             & 0.6785 & 0.7382 & 0.1230 \\
        RACE--MMLU\&SWAG             & 0.5855 & 0.7267 & 0.1419 \\
        SWAG--MMLU\&RACE             & 0.5875 & 0.6821 & 0.1046 \\
        \bottomrule
    \end{tabular}
    \label{tab:metrics_setting2}
\end{table}

\begin{table}[H]
    \centering
    \caption{Poor cross-task generalisation performance using only 1-layer-hidden-state features on short-form question answering datasets using Mistral-7B as the LLM.}
    \begin{tabular}{lccc}
        \toprule
        \textbf{Dataset(s)} & \textbf{Acc} & \textbf{ROC AUC} & \textbf{ECE} \\
        \midrule
        TriviaQA                          & 0.7445 & 0.8201 & 0.1405 \\
        Squad                             & 0.7392 & 0.8280 & 0.1287 \\
        Winogrande                        & 0.7400 & 0.8549 & 0.1133 \\
        TriviaQA--Squad                   & 0.6440 & 0.7175 & 0.1179 \\
        TriviaQA--Winogrande              & 0.5015 & 0.5917 & 0.0857 \\
        Squad--TriviaQA                   & 0.6379 & 0.7873 & 0.1742 \\
        Squad--Winogrande                 & 0.5475 & 0.6486 & 0.0706 \\
        Winogrande--TriviaQA              & 0.5000 & 0.5831 & 0.1218 \\
        Winogrande--Squad                 & 0.5735 & 0.6234 & 0.0740 \\
        TriviaQA--Squad\&Winogrande       & 0.6285 & 0.7071 & 0.1053 \\
        Squad--TriviaQA\&Winogrande       & 0.7038 & 0.7797 & 0.1649 \\
        Winogrande--TriviaQA\&Squad       & 0.5360 & 0.5879 & 0.0571 \\
        \bottomrule
    \end{tabular}
    \label{tab:triviaqa_squad_winogrande_2}
\end{table}

As we can see from Tables 1 to 4, probes trained on one dataset generalise poorly when tested on other datasets. Specifically, using only hidden-state features from LLaMA2-7B and Mistral-7B exhibit significant drops in performance, especially in accuracy, when evaluated in cross-task settings. \\\\
These results demonstrate that hidden-state-based uncertainty estimators trained in a single-task setting do not generalise well to new tasks or datasets. This suggests that such probes may be overfitting to dataset-specific features rather than capturing generalisable representations. \\

\textbf{Effect of data-agnostic features on the generalisation performance of the trained probe \footnote{Refer to the Complementary Document in our GitHub repository for additional experimental results on this section.}}

\begin{table}[H]
  \centering
  \caption{Generalisation performance between not including data-agnostic features (4096 features) and including data-agnostic features (4096 + 5 = 4101 features) on multiple-choice question answering datasets using Llama2--7B as the LLM.}
  \label{tab:generalisation-performance}
  \setlength{\tabcolsep}{4pt}
  \begin{adjustbox}{max width=\textwidth}
    \begin{tabular}{l  ccc  ccc}
      \toprule
      \multirow{2}{*}{\textbf{Transfer pairs}}
        & \multicolumn{3}{c}{\textbf{No data-agnostic features}}
        & \multicolumn{3}{c}{\textbf{With 5 data-agnostic features}} \\
      \cmidrule(lr){2-4} \cmidrule(lr){5-7}
        & \textbf{Acc} & \textbf{ROC AUC} & \textbf{ECE}
        & \textbf{Acc} & \textbf{ROC AUC} & \textbf{ECE} \\
      \midrule
      MMLU--RACE         & 0.6095 & 0.6629 & \textbf{0.0761} & \textbf{0.6355} & \textbf{0.6807} & 0.1007 \\
      RACE--MMLU         & 0.5960 & 0.6714 & \textbf{0.0940} & \textbf{0.6155} & \textbf{0.6975} & 0.1023 \\
      RACE--SWAG         & 0.5160 & 0.5197 & \textbf{0.0144} & \textbf{0.5310} & \textbf{0.5529} & 0.0194 \\
      SWAG--RACE         & \textbf{0.5295} & \textbf{0.5339} & \textbf{0.0221} & 0.5075 & 0.5139 & 0.0282 \\
      MMLU--SWAG         & \textbf{0.5170} & \textbf{0.5473} & 0.0155 & \textbf{0.5170} & 0.5223 & \textbf{0.0131} \\
      SWAG--MMLU         & 0.5255 & \textbf{0.5472} & 0.0128 & \textbf{0.5270} & 0.5422 & \textbf{0.0095} \\
      MMLU--SWAG\&RACE   & 0.6295 & 0.6815 & \textbf{0.0890} & \textbf{0.6440} & \textbf{0.6896} & 0.1007 \\
      RACE--MMLU\&SWAG   & 0.5900 & 0.6839 & \textbf{0.1136} & \textbf{0.6110} & \textbf{0.7085} & 0.1283 \\
      SWAG--MMLU\&RACE   & 0.5270 & \textbf{0.5476} & 0.0303 & \textbf{0.5315} & 0.5420 & \textbf{0.0084} \\
      \bottomrule
    \end{tabular}
  \end{adjustbox}
\end{table}

\begin{table}[H]
  \centering
  \caption{Generalisation performance between not including data‐agnostic features (4096 features) and including data‐agnostic features (4096 + 5 = 4101 features) on multiple‐choice question answering datasets using Mistral--7B as the LLM.}
  \label{tab:generalisation-performance-mistral}
  \setlength{\tabcolsep}{4pt}
  \begin{adjustbox}{max width=\textwidth}
    \begin{tabular}{l  ccc  ccc}
      \toprule
      \multirow{2}{*}{\textbf{Transfer pairs}}
        & \multicolumn{3}{c}{\textbf{No data‐agnostic features}}
        & \multicolumn{3}{c}{\textbf{With 5 data‐agnostic features}} \\
      \cmidrule(lr){2-4} \cmidrule(lr){5-7}
        & \textbf{Acc} & \textbf{ROC AUC} & \textbf{ECE}
        & \textbf{Acc} & \textbf{ROC AUC} & \textbf{ECE} \\
      \midrule
      MMLU--RACE       & 0.6725 & 0.7349 & \textbf{0.1284} & \textbf{0.6820} & \textbf{0.7568} & 0.1397 \\
      RACE--MMLU       & 0.5870 & 0.7307 & \textbf{0.1390} & \textbf{0.6185} & \textbf{0.7573} & 0.1468 \\
      RACE--SWAG       & 0.5400 & 0.6401 & \textbf{0.0772} & \textbf{0.5690} & \textbf{0.6763} & 0.1044 \\
      SWAG--RACE       & 0.5405 & 0.6494 & \textbf{0.0998} & \textbf{0.5410} & \textbf{0.6693} & 0.1282 \\
      MMLU--SWAG       & 0.5700 & 0.6397 & \textbf{0.0754} & \textbf{0.5860} & \textbf{0.6816} & 0.1162 \\
      SWAG--MMLU       & 0.6145 & 0.6635 & \textbf{0.0877} & \textbf{0.6230} & \textbf{0.6786} & 0.0970 \\
      MMLU--SWAG\&RACE & 0.6785 & 0.7382 & \textbf{0.1230} & \textbf{0.7065} & \textbf{0.7601} & 0.1357 \\
      RACE--MMLU\&SWAG & 0.5855 & 0.7267 & \textbf{0.1419} & \textbf{0.6055} & \textbf{0.7527} & 0.1518 \\
      SWAG--MMLU\&RACE & \textbf{0.5875} & 0.6821 & \textbf{0.1046} & 0.5770 & \textbf{0.6946} & 0.1251 \\
      \bottomrule
    \end{tabular}
  \end{adjustbox}
\end{table}

\begin{table}[H]
  \centering
  \caption{Generalisation performance between not including data‐agnostic features (4096 features) and including data‐agnostic features (4096 + 4 = 4100 features) on short‐form question answering datasets using Llama2--7B as the LLM.}
  \label{tab:generalisation-performance-shortform}
  \setlength{\tabcolsep}{4pt}
  \begin{adjustbox}{max width=\textwidth}
    \begin{tabular}{l  ccc  ccc}
      \toprule
      \multirow{2}{*}{\textbf{Transfer pairs}}
        & \multicolumn{3}{c}{\textbf{No data‐agnostic features}}
        & \multicolumn{3}{c}{\textbf{With 4 data‐agnostic features}} \\
      \cmidrule(lr){2-4} \cmidrule(lr){5-7}
        & \textbf{Acc} & \textbf{ROC AUC} & \textbf{ECE}
        & \textbf{Acc} & \textbf{ROC AUC} & \textbf{ECE} \\
      \midrule
      TriviaQA--Squad             & 0.6350 & 0.6690 & \textbf{0.0822} & \textbf{0.7155} & \textbf{0.7819} & 0.1537 \\
      TriviaQA--Winogrande        & 0.5135 & 0.5646 & \textbf{0.0427} & \textbf{0.5435} & \textbf{0.6342} & 0.0715 \\
      Squad--TriviaQA             & 0.5959 & 0.7328 & \textbf{0.1428} & \textbf{0.7087} & \textbf{0.7888} & 0.1590 \\
      Squad--Winogrande           & 0.5221 & 0.5705 & \textbf{0.0322} & \textbf{0.5343} & \textbf{0.5976} & 0.0453 \\
      Winogrande--TriviaQA        & 0.5020 & 0.5489 & \textbf{0.0891} & \textbf{0.5390} & \textbf{0.5871} & 0.0791 \\
      Winogrande--Squad           & 0.5310 & 0.5529 & \textbf{0.0170} & \textbf{0.5540} & \textbf{0.5843} & 0.0251 \\
      TriviaQA--Squad\&Winogrande & 0.6170 & 0.6587 & \textbf{0.0750} & \textbf{0.7060} & \textbf{0.7764} & 0.1554 \\
      Squad--TriviaQA\&Winogrande & 0.6064 & 0.7352 & \textbf{0.1424} & \textbf{0.7093} & \textbf{0.7814} & 0.1573 \\
      Winogrande--TriviaQA\&Squad  & 0.5320 & 0.5688 & \textbf{0.0394} & \textbf{0.5500} & \textbf{0.5928} & 0.0482 \\
      \bottomrule
    \end{tabular}
  \end{adjustbox}
\end{table}

\begin{table}[H]
  \centering
  \caption{Generalisation performance between not including data‐agnostic features (4096 features) and including data‐agnostic features (4096 + 4 = 4100 features) on short‐form question answering datasets using Mistral--7B as the LLM.}
  \label{tab:generalisation-performance-shortform-mistral}
  \setlength{\tabcolsep}{4pt}
  \begin{adjustbox}{max width=\textwidth}
    \begin{tabular}{l  ccc  ccc}
      \toprule
      \multirow{2}{*}{\textbf{Transfer pairs}}
        & \multicolumn{3}{c}{\textbf{No data‐agnostic features}}
        & \multicolumn{3}{c}{\textbf{With 4 data‐agnostic features}} \\
      \cmidrule(lr){2-4} \cmidrule(lr){5-7}
        & \textbf{Acc} & \textbf{ROC AUC} & \textbf{ECE}
        & \textbf{Acc} & \textbf{ROC AUC} & \textbf{ECE} \\
      \midrule
      TriviaQA--Squad             & 0.6440 & 0.7175 & \textbf{0.1179} & \textbf{0.6785} & \textbf{0.7804} & 0.1667 \\
      TriviaQA--Winogrande        & 0.5015 & 0.5917 & \textbf{0.0857} & \textbf{0.5090} & \textbf{0.6092} & 0.0956 \\
      Squad--TriviaQA             & 0.6379 & 0.7873 & \textbf{0.1742} & \textbf{0.7361} & \textbf{0.8260} & 0.1838 \\
      Squad--Winogrande           & 0.5475 & 0.6486 & \textbf{0.0706} & \textbf{0.6033} & \textbf{0.6810} & 0.1072 \\
      Winogrande--TriviaQA        & 0.5000 & 0.5831 & \textbf{0.1218} & \textbf{0.5590} & \textbf{0.6413} & 0.0804 \\
      Winogrande--Squad           & 0.5735 & 0.6234 & \textbf{0.0740} & \textbf{0.5870} & \textbf{0.6398} & 0.0661 \\
      TriviaQA--Squad\&Winogrande & 0.6285 & 0.7071 & \textbf{0.1053} & \textbf{0.6615} & \textbf{0.7640} & 0.1572 \\
      Squad--TriviaQA\&Winogrande & 0.7038 & 0.7797 & \textbf{0.1649} & \textbf{0.7196} & \textbf{0.8086} & 0.1663 \\
      Winogrande--TriviaQA\&Squad  & 0.5360 & 0.5879 & \textbf{0.0571} & \textbf{0.5950} & \textbf{0.6339} & 0.0574 \\
      \bottomrule
    \end{tabular}
  \end{adjustbox}
\end{table}

We can see from Tables 5 to 8 that incorporating data-agnostic features into the set of features to train the probe helps improve its generalisation performance, i.e., 7 improved out of 9 for Llama2 – 7B on multiple choice question answering datasets; 8 improved out of 9 for Mistral – 7B on short-form question answering datasets; 9 improved out of 9 for Llama2 – 7B on multiple choice question answering datasets; and 9 improved out of 9 for Mistral – 7B on short-form question answering datasets. Additionally, incorporating data-agnostic features into the feature set for short-form question answering datasets yields larger improvement in generalisation compared to multiple-choice question answering datasets. \\

In addition, when we observe the results more closely, we notice an asymmetric generalisation pattern. For example, incorporating data-agnostic features helps improve the generalisation performance in “RACE-SWAG” case but fails in “SWAG-RACE” case. A similar pattern was also reported in \cite{Orgad2024intrinsic}, which remains unexplained. \\

\textbf{Generalisation performance improvement when including data-agnostic features under three configurations – top-k selected features, 1 middle layer, and 5 middle layers \footnote{Refer to the Complementary Document in our GitHub repository for additional experimental results on this section.}}

\begin{figure}[H]
  \centering
  \includegraphics[width=0.9\textwidth]{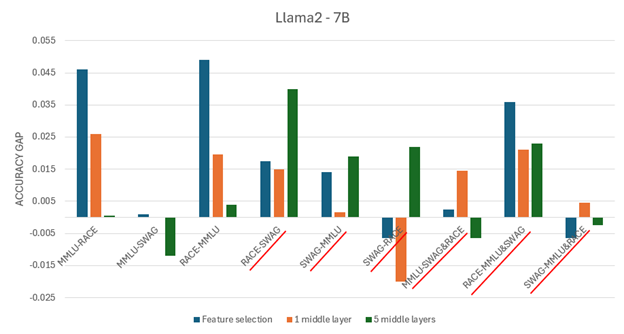}
  \caption{Accuracy gap between not including data-agnostic features and including data-agnostic features under three configurations in multiple-choice question answering datasets using Llama2–7B as the LLM.}
  \label{fig:llama2-accuracy-mc-gap}
\end{figure}

\begin{figure}[H]
  \centering
  \includegraphics[width=0.9\textwidth]{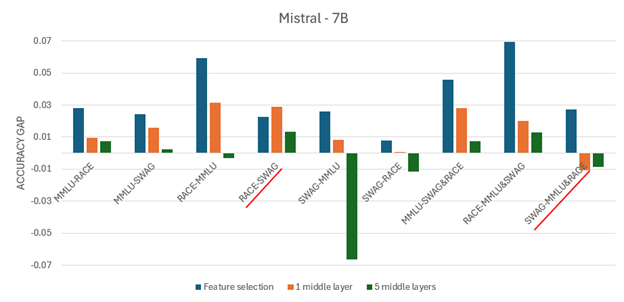}
  \caption{Accuracy gap between not including data-agnostic features and including data-agnostic features under three configurations in multiple-choice question answering datasets using Mistral – 7B as the LLM.}
  \label{fig:mistral-mc-accuracy-gap}
\end{figure}

\begin{figure}[H]
  \centering
  \includegraphics[width=0.9\textwidth]{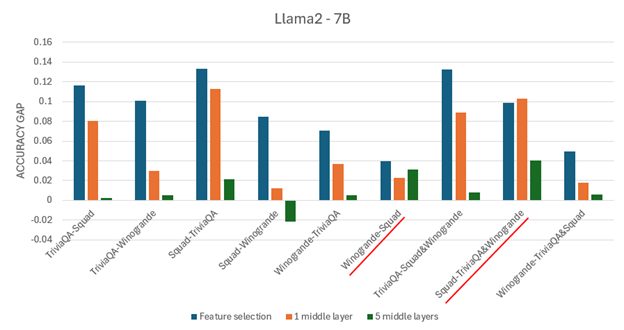}
  \caption{Accuracy gap between not including data-agnostic features and including data-agnostic features under three configurations in short-form question answering datasets using Llama2 – 7B as the LLM.}
  \label{fig:llama2-accuracy-sf-gap}
\end{figure}

\begin{figure}[H]
  \centering
  \includegraphics[width=0.9\textwidth]{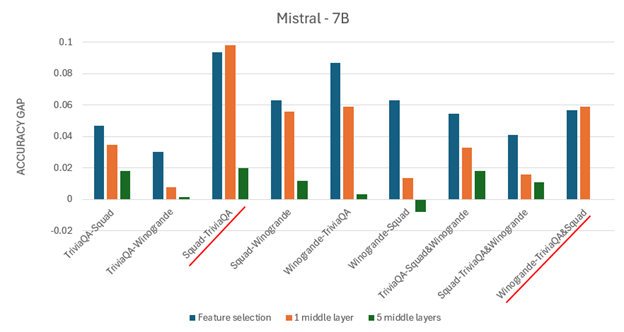}
  \caption{Accuracy gap between not including data-agnostic features and including data-agnostic features under three configurations in short-form question answering datasets using Mistral – 7B as the LLM.}
  \label{fig:mistral-accuracy-sf-gap}
\end{figure}

As mentioned in the Methodology section, the configuration that has the least hidden‐state features (but still be able to preserve the information by applying feature selection using Pearson Correlation Coefficient with the ground truth labels) will yield the largest improvement when incorporating data‐agnostic features as the task‐specific features are reduced, helping amplify the effect of the data‐agnostic features. We therefore expected the selected top‐k setting to give the largest boost, the 1‐layer setting the next largest, and the 5‐layer setting the smallest. \\

However, Figures 1 to 4 show that this ordering does not always hold. The pattern appears in only 3 out of 9 transfer pairs for Llama2 – 7B on multiple‐choice question answering datasets, but in 7 out of 9 pairs for each of the other three model–dataset type combinations. To investigate what the potential reasons could be, we conduct an additional experiment focusing on feature contributions, which will be discussed in the next section. \\

\textbf{Utilising SHAP values to investigate feature contributions \footnote{Refer to the Excel Files folder in our GitHub repository for the remaining experimental results on this section.}}\\

To investigate feature contributions, especially focusing on the data‐agnostic features, we utilise Tree SHAP method \citep{Lundberg2017shap} to obtain the contribution of each feature to the model’s predictions. Specifically, for \(n\) data points within a dataset (we use the test set), we will have the corresponding \(n\) absolute SHAP values, which will then be averaged to get the mean SHAP value. The higher the SHAP value, the larger the contribution to the predictions.

We label each SHAP table as \verb|<LLM> - <Dataset> - <Feature set>| where:
\verb|<LLM>| refers to the language model that produced the hidden states (e.g., Llama2-7B or Mistral-7B). 
\verb|Dataset| denotes the benchmark whose training split is used to fit a random forest regressor and whose test split is used to compute SHAP values.
\verb|Feature set| indicates one of three hidden state configurations: 
(1) \textit{Feature selection}, consisting of 305 features for multiple-choice question answering datasets and 304 features for short-form question answering datasets; 
(2) \textit{1 middle layer}, comprising 4101 features for multiple-choice and 4100 for short-form datasets; and 
(3) \textit{5 middle layers}, with 20485 features for multiple-choice and 20484 for short-form datasets.
\\

In every case, the input to the regressor contains both the chosen hidden state features and the data-agnostic features; the mean SHAP values quantify how much each feature contributes on the test split. \\

For most cases, we will have the following observations (taking Mistral – MMLU as an example):
\begin{enumerate}
  \item First, data-agnostic features (highlighted in \textbf{bold}) have the highest SHAP values compared to other features which are the hidden units from the LLM.
  \item Second, under three cases – 305 features (utilise feature selection), 4096 features (1 middle layer) and 20485 features (5 middle layers), the data-agnostic features in the “305 features” setting have the highest SHAP values, followed by the “4101 features” setting and finally the “20485 features” setting. Specifically in the example below, the data-agnostic features in the “305 features” setting have SHAP values in the range of (0.018, 0.052), for the “4101 features” setting, the range is (0.008, 0.011), and for the “20485 features” setting, the range is (0.001, 0.002).
\end{enumerate}

\begin{table}[H]
  \centering
  \caption{Mistral - MMLU - 305 features}
  \label{tab:shap-mistral-mmlu-305}
  \setlength{\tabcolsep}{4pt}
  \begin{adjustbox}{max width=\textwidth}
    \begin{tabular}{c l r}
      \toprule
      \#  & \textbf{Feature}   & \textbf{Mean SHAP} \\
      \midrule
      0  & \textbf{feature\_300}  & 0.051938013 \\
      1  & \textbf{feature\_304}  & 0.039524272 \\
      2  & \textbf{feature\_301}  & 0.035966115 \\
      3  & \textbf{feature\_302}  & 0.026267883 \\
      4  & \textbf{feature\_303}  & 0.018435749 \\
      5  & feature\_16   & 0.004894765 \\
      6  & feature\_10   & 0.004719694 \\
      7  & feature\_1    & 0.004630801 \\
      8  & feature\_23   & 0.004069274 \\
      9  & feature\_2    & 0.003971775 \\
      \bottomrule
    \end{tabular}
  \end{adjustbox}
\end{table}

\begin{table}[H]
  \centering
  \caption{Mistral - MMLU - 4101 features}
  \label{tab:shap-mistral-mmlu-4101}
  \setlength{\tabcolsep}{4pt}
  \begin{adjustbox}{max width=\textwidth}
    \begin{tabular}{c l r}
      \toprule
      \# & \textbf{Feature}   & \textbf{Mean SHAP} \\
      \midrule
      0 & \textbf{feature\_4100} & 0.011295522 \\
      1 & \textbf{feature\_4097} & 0.010986307 \\
      2 & \textbf{feature\_4096} & 0.010587706 \\
      3 & \textbf{feature\_4099} & 0.009048645 \\
      4 & \textbf{feature\_4098} & 0.008210684 \\
      5 & feature\_2100 & 0.003257098 \\
      6 & feature\_1888 & 0.002641738 \\
      7 & feature\_2614 & 0.002445704 \\
      8 & feature\_2043 & 0.002435461 \\
      \bottomrule
    \end{tabular}
  \end{adjustbox}
\end{table}

\begin{table}[H]
  \centering
  \caption{Mistral - MMLU - 20485 features}
  \label{tab:shap-mistral-mmlu-20485}
  \setlength{\tabcolsep}{4pt}
  \begin{adjustbox}{max width=\textwidth}
    \begin{tabular}{c l r}
      \toprule
      \# & \textbf{Feature}   & \textbf{Mean SHAP} \\
      \midrule
      0 & \textbf{feature\_20480} & 0.002624 \\
      1 & \textbf{feature\_20481} & 0.002535 \\
      2 & \textbf{feature\_20484} & 0.002478 \\
      3 & \textbf{feature\_20483} & 0.001850 \\
      4 & feature\_19014 & 0.001493 \\
      5 & \textbf{feature\_20482} & 0.001448 \\
      6 & feature\_4110  & 0.001296 \\
      7 & feature\_12000 & 0.001139 \\
      8 & feature\_14    & 0.001110 \\
      9 & feature\_13618 & 0.001032 \\
      \bottomrule
    \end{tabular}
  \end{adjustbox}
\end{table}

These results imply that pruning task‑specific hidden units amplifies the relative importance of the data‑agnostic features. \\

However, there also exist some cases where the data-agnostic features are not ranked at the top or sparsely distributed in ranking: \\

The first and the most notable one is \textbf{Llama – SWAG} where for all three settings, all of the data-agnostic features do not have the highest mean SHAP values, this means that the data-agnostic features do not have the largest contribution in this case. Since for the other two datasets – such as MMLU and RACE – where data‑agnostic features consistently rank at the top, the cross-dataset evaluation involving SWAG dataset might affect the expected ordering we have for RQ2. \\

The second is \textbf{Mistral – SWAG}, illustrated in Tables 12 to 14.

\begin{table}[H]
  \centering
  \caption{Mistral - SWAG - 305 features}
  \label{tab:shap-mistral-swag-305}
  \setlength{\tabcolsep}{4pt}
  \begin{adjustbox}{max width=\textwidth}
    \begin{tabular}{c l r}
      \toprule
      \# & \textbf{Feature}      & \textbf{Mean SHAP} \\
      \midrule
      0  & \textbf{feature\_300} & 0.031627626 \\
      1  & feature\_0   & 0.028910070 \\
      2  & \textbf{feature\_301} & 0.024511878 \\
      3  & \textbf{feature\_304} & 0.019163538 \\
      4  & feature\_2   & 0.016880925 \\
      5  & feature\_1   & 0.014940283 \\
      6  & \textbf{feature\_302} & 0.013342291 \\
      7  & feature\_26  & 0.013170929 \\
      8  & feature\_5   & 0.010116517 \\
      9  & feature\_8   & 0.009928698 \\
      \midrule
      \multicolumn{3}{c}{\dots} \\
      83 & \textbf{feature\_303} & 0.001781455 \\
      \bottomrule
    \end{tabular}
  \end{adjustbox}
\end{table}

\begin{table}[H]
  \centering
  \caption{Mistral - SWAG - 4101 features}
  \label{tab:shap-mistral-swag-4101}
  \setlength{\tabcolsep}{4pt}
  \begin{adjustbox}{max width=\textwidth}
    \begin{tabular}{c l r}
      \toprule
      \#  & \textbf{Feature}    & \textbf{Mean SHAP} \\
      \midrule
      0  & \textbf{feature\_4097} & 0.006342038 \\
      1  & feature\_1680 & 0.006241095 \\
      2  & \textbf{feature\_4096} & 0.005953367 \\
      3  & \textbf{feature\_4098} & 0.005877953 \\
      4  & feature\_3440 & 0.004557681 \\
      5  & \textbf{feature\_4100} & 0.004414030 \\
      6  & feature\_3049 & 0.004042185 \\
      7  & feature\_315  & 0.003773831 \\
      8  & feature\_1785 & 0.003502831 \\
      9  & feature\_534  & 0.003502057 \\
      \midrule
      \multicolumn{3}{c}{\dots} \\
      19 & \textbf{feature\_4099} & 0.002680176 \\
      \bottomrule
    \end{tabular}
  \end{adjustbox}
\end{table}

\begin{table}[H]
  \centering
  \caption{Mistral - SWAG – 20485 features}
  \label{tab:shap-mistral-swag-20485}
  \setlength{\tabcolsep}{4pt}
  \begin{adjustbox}{max width=\textwidth}
    \begin{tabular}{c l r}
      \toprule
      \#   & \textbf{Feature}    & \textbf{Mean SHAP} \\
      \midrule
      0    & feature\_15594      & 0.001714757 \\
      1    & \textbf{feature\_20481}      & 0.001650367 \\
      2    & feature\_9872       & 0.001475606 \\
      3    & feature\_10921      & 0.001346881 \\
      4    & \textbf{feature\_20480}      & 0.001327448 \\
      5    & feature\_7085       & 0.001203152 \\
      6    & \textbf{feature\_20484}      & 0.001194739 \\
      7    & feature\_4073       & 0.001178399 \\
      8    & feature\_20104      & 0.001142685 \\
      9    & feature\_4649       & 0.001087810 \\
      \midrule
      \multicolumn{3}{c}{\dots} \\
      66   & \textbf{feature\_20482}      & 0.000689289 \\
      \multicolumn{3}{c}{\dots} \\
      72   & \textbf{feature\_20483}      & 0.000668030 \\
      \bottomrule
    \end{tabular}
  \end{adjustbox}
\end{table}

The third is \textbf{Llama – Winogrande}, illustrated in Tables 15 to 17.

\begin{table}[H]
  \centering
  \caption{Llama - Winogrande - 304 features}
  \label{tab:shap-llama-winogrande-304}
  \setlength{\tabcolsep}{4pt}
  \begin{adjustbox}{max width=\textwidth}
    \begin{tabular}{c l r}
      \toprule
      \#  & \textbf{Feature}   & \textbf{Mean SHAP} \\
      \midrule
      0  & \textbf{feature\_301} & 0.016738563 \\
      1  & \textbf{feature\_300} & 0.016018875 \\
      2  & feature\_105 & 0.005134809 \\
      3  & feature\_15  & 0.004811519 \\
      4  & feature\_241 & 0.004561076 \\
      5  & feature\_200 & 0.003893218 \\
      6  & \textbf{feature\_303} & 0.003729269 \\
      7  & feature\_58  & 0.003683539 \\
      8  & feature\_30  & 0.003678206 \\
      9  & feature\_43  & 0.003469689 \\
      \midrule
      \multicolumn{3}{c}{\dots} \\
      43 & \textbf{feature\_302} & 0.001838276 \\
      \bottomrule
    \end{tabular}
  \end{adjustbox}
\end{table}

\begin{table}[H]
  \centering
  \caption{Llama - Winogrande - 4100 features}
  \label{tab:shap-llama-winogrande-4100}
  \setlength{\tabcolsep}{4pt}
  \begin{adjustbox}{max width=\textwidth}
    \begin{tabular}{c l r}
      \toprule
      \#  & \textbf{Feature}    & \textbf{Mean SHAP} \\
      \midrule
      0  & \textbf{feature\_4097} & 0.003184 \\
      1  & \textbf{feature\_4096} & 0.002803 \\
      2  & \textbf{feature\_4098} & 0.001437 \\
      3  & \textbf{feature\_4099} & 0.001071 \\
      4  & feature\_583  & 0.001066 \\
      5  & feature\_2298 & 0.001059 \\
      6  & feature\_1924 & 0.000260 \\
      7  & feature\_1233 & 0.000092 \\
      8  & feature\_2172 & 0.000894 \\
      \bottomrule
    \end{tabular}
  \end{adjustbox}
\end{table}

\begin{table}[H]
  \centering
  \caption{Llama - Winogrande - 20484 features}
  \label{tab:shap-llama-winogrande-20484}
  \setlength{\tabcolsep}{4pt}
  \begin{adjustbox}{max width=\textwidth}
    \begin{tabular}{c l r}
      \toprule
      \#   & \textbf{Feature}    & \textbf{Mean SHAP} \\
      \midrule
      0    & \textbf{feature\_20480}      & 0.000704278 \\
      1    & \textbf{feature\_20481}      & 0.000567875 \\
      2    & feature\_16554      & 0.000556392 \\
      3    & feature\_12980      & 0.000511399 \\
      4    & feature\_11331      & 0.000487444 \\
      5    & feature\_17783      & 0.000485209 \\
      6    & feature\_10913      & 0.000476228 \\
      7    & feature\_14101      & 0.000458615 \\
      8    & feature\_15300      & 0.000454917 \\
      9    & feature\_5836       & 0.000450639 \\
      \midrule
      \multicolumn{3}{c}{\dots} \\
      188  & \textbf{feature\_20482}      & 0.000238807 \\
      \multicolumn{3}{c}{\dots} \\
      191  & \textbf{feature\_20483}      & 0.000237239 \\
      \bottomrule
    \end{tabular}
  \end{adjustbox}
\end{table}

The fourth is \textbf{Mistral – Winogrande}, illustrated in Tables 18 to 20.

\begin{table}[H]
  \centering
  \caption{Mistral - Winogrande - 304 features}
  \label{tab:shap-mistral-winogrande-304}
  \setlength{\tabcolsep}{4pt}
  \begin{adjustbox}{max width=\textwidth}
    \begin{tabular}{c l r}
      \toprule
      \#  & \textbf{Feature}   & \textbf{Mean SHAP} \\
      \midrule
      0  & \textbf{feature\_301} & 0.041023226 \\
      1  & \textbf{feature\_300} & 0.020536170 \\
      2  & \textbf{feature\_303} & 0.016145899 \\
      3  & feature\_155 & 0.006490395 \\
      4  & feature\_22  & 0.005999329 \\
      5  & feature\_67  & 0.005727566 \\
      6  & feature\_0   & 0.005684195 \\
      7  & feature\_101 & 0.005676837 \\
      8  & feature\_36  & 0.005562444 \\
      9  & feature\_178 & 0.005117074 \\
      10 & \textbf{feature\_302} & 0.004537059 \\
      \bottomrule
    \end{tabular}
  \end{adjustbox}
\end{table}

\begin{table}[H]
  \centering
  \caption{Mistral - Winogrande - 4100 features}
  \label{tab:shap-mistral-winogrande-4100}
  \setlength{\tabcolsep}{4pt}
  \begin{adjustbox}{max width=\textwidth}
    \begin{tabular}{c l r}
      \toprule
      \#  & \textbf{Feature}    & \textbf{Mean SHAP} \\
      \midrule
      0  & \textbf{feature\_4097} & 0.006562 \\
      1  & \textbf{feature\_4096} & 0.004854 \\
      2  & \textbf{feature\_4099} & 0.003801 \\
      3  & \textbf{feature\_4098} & 0.002216 \\
      4  & feature\_710  & 0.001941 \\
      5  & feature\_492  & 0.001731 \\
      6  & feature\_2082 & 0.001581 \\
      7  & feature\_3989 & 0.001533 \\
      8  & feature\_3681 & 0.001507 \\
      9  & feature\_3761 & 0.001502 \\
      \bottomrule
    \end{tabular}
  \end{adjustbox}
\end{table}

\begin{table}[H]
  \centering
  \caption{Mistral - Winogrande - 20484 features}
  \label{tab:shap-mistral-winogrande-20484}
  \setlength{\tabcolsep}{4pt}
  \begin{adjustbox}{max width=\textwidth}
    \begin{tabular}{c l r}
      \toprule
      \#   & \textbf{Feature}    & \textbf{Mean SHAP} \\
      \midrule
      0    & \textbf{feature\_20481}      & 0.001276846 \\
      1    & feature\_18828      & 0.000851321 \\
      2    & feature\_18628      & 0.000827854 \\
      3    & feature\_16731      & 0.000819973 \\
      4    & feature\_12853      & 0.000807152 \\
      5    & \textbf{feature\_20480}      & 0.000805161 \\
      6    & feature\_4540       & 0.000739713 \\
      7    & \textbf{feature\_20482}      & 0.000728036 \\
      8    & feature\_16911      & 0.000680573 \\
      9    & feature\_19303      & 0.000651299 \\
      \midrule
      \multicolumn{3}{c}{\dots} \\
      402  & \textbf{feature\_20483}      & 0.000224495 \\
      \bottomrule
    \end{tabular}
  \end{adjustbox}
\end{table}

For the three outlier cases, Llama – Winogrande, Mistral – SWAG, Mistral – Winogrande, the data-agnostic features rank sparsely, with some being ranked at the top while the remaining being ranked lower. However, it is not as critical as the Llama – SWAG case as the selected top-k and 1 middle layer settings still have most of the data-agnostic features being ranked at the top, hence, the expected ordering holds in most of the cases.

\section{Ablation Studies}

\subsection{The number of correct answers misclassified and the number of new correct answers when incorporating data-agnostic features} 

To further investigate the effectiveness of the data-agnostic features in improving performance, we calculate the number of correct answers when not incorporating data-agnostic features being misclassified when incorporating data-agnostic features and the number of new correct answers when incorporating data-agnostic features. Technically, we denote the list of indices that have correct answers when not incorporating data-agnostic features as L1 and denote the list of indices that have correct answers when incorporating data-agnostic features as L2. Then, the two metrics can be calculated as follows: 
\begin{align*}
\text{Number of correct answers misclassified} &= |L1 \setminus L2| \\
\text{Number of new correct answers} &= |L2 \setminus L1|
\end{align*}

In general, the number of correct answers being misclassified when incorporating data-agnostic features is relatively high for most of the cases. Whether the accuracy when incorporating data-agnostic features is higher than that when not incorporating data-agnostic features or not depends on the number of new correct answers. Given that the number of correct answers that turned incorrect is high, the number of new correct answers must be higher for the accuracy when incorporating data-agnostic features to be higher. An example is presented in Table 21 \footnote{Refer to the Complementary Document in our GitHub repository for full experimental results on this section.}.

\begin{table}[H]
  \centering
  \caption{Changes in classification outcomes when adding data-agnostic features.}
  \label{tab:classification-changes}
  \setlength{\tabcolsep}{4pt}
  \begin{tabular}{llr}
    \toprule
    \textbf{Transfer pairs} & \textbf{Metric}                     & \textbf{Count} \\
    \midrule
    \multirow{2}{*}{MMLU--RACE}       & Correct turned incorrect & 199 \\
                                      & New correct             & 251 \\
    \midrule
    \multirow{2}{*}{RACE--MMLU}       & Correct turned incorrect & 164 \\
                                      & New correct             & 203 \\
    \midrule
    \multirow{2}{*}{RACE--SWAG}       & Correct turned incorrect & 443 \\
                                      & New correct             & 473 \\
    \midrule
    \multirow{2}{*}{SWAG--RACE}       & Correct turned incorrect & 403 \\
                                      & New correct             & 359 \\
    \midrule
    \multirow{2}{*}{MMLU--SWAG}       & Correct turned incorrect & 425 \\
                                      & New correct             & 425 \\
    \midrule
    \multirow{2}{*}{SWAG--MMLU}       & Correct turned incorrect & 403 \\
                                      & New correct             & 406 \\
    \midrule
    \multirow{2}{*}{MMLU--SWAG\&RACE} & Correct turned incorrect & 182 \\
                                      & New correct             & 211 \\
    \midrule
    \multirow{2}{*}{RACE--MMLU\&SWAG} & Correct turned incorrect & 121 \\
                                      & New correct             & 163 \\
    \midrule
    \multirow{2}{*}{SWAG--MMLU\&RACE} & Correct turned incorrect & 365 \\
                                      & New correct             & 374 \\
    \bottomrule
  \end{tabular}
\end{table}

\subsection{Factual QA task might share some similarities with Reading Comprehension task}

Upon a close look at the results in Tables 5 to 8 in the “Not including data-agnostic features” column, we notice that for every cross-dataset case that involves a dataset from Factual QA task and a dataset from Reading Comprehension task, the generalisation performance is higher compared to the combinations involving a dataset from the Commonsense Reasoning task. We utilise PCA to observe the distribution of the hidden-state features among these three tasks. As we can see in Figure 5 and 6, the clusters of Factual QA datasets and Reading Compreshension datasets are closer together. This implies that Factual QA and Reading Compreshension tasks might share some similarities, which results in better generalisation performance. A similar pattern has also been mentioned in \cite{Orgad2024intrinsic} in which they stated that generalisation occurs within tasks requiring similar skills.

\begin{figure}[H]
  \centering
  \begin{subfigure}[b]{0.48\textwidth}
    \centering
    \includegraphics[width=\textwidth]{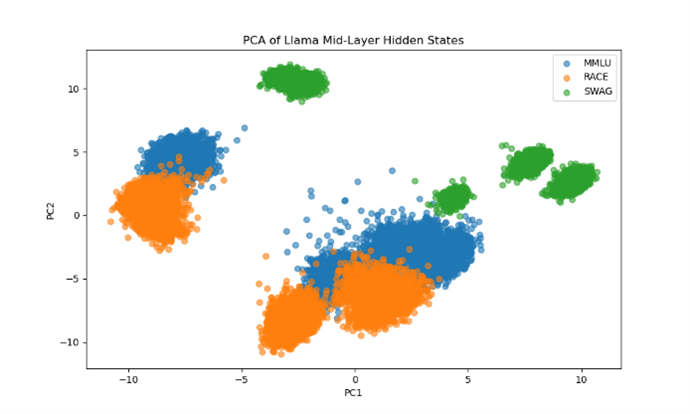}
    \caption{Multiple-choice QA}
    \label{fig:pca-mcqa}
  \end{subfigure}
  \hfill
  \begin{subfigure}[b]{0.48\textwidth}
    \centering
    \includegraphics[width=\textwidth]{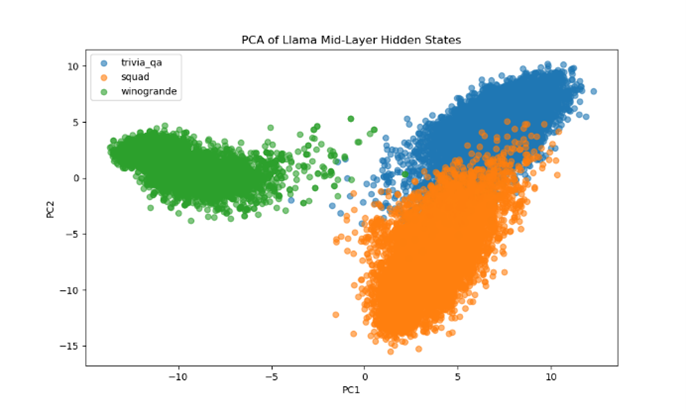}
    \caption{Short-form QA}
    \label{fig:pca-sfqa}
  \end{subfigure}

  \caption{PCA of the hidden states for multiple choice question answering (left) and short-form question answering datasets (right) using Llama2 – 7B as the LLM.}
  \label{fig:pca-llama2}
\end{figure}

\begin{figure}[H]
  \centering
  \begin{subfigure}[b]{0.48\textwidth}
    \centering
    \includegraphics[width=\textwidth]{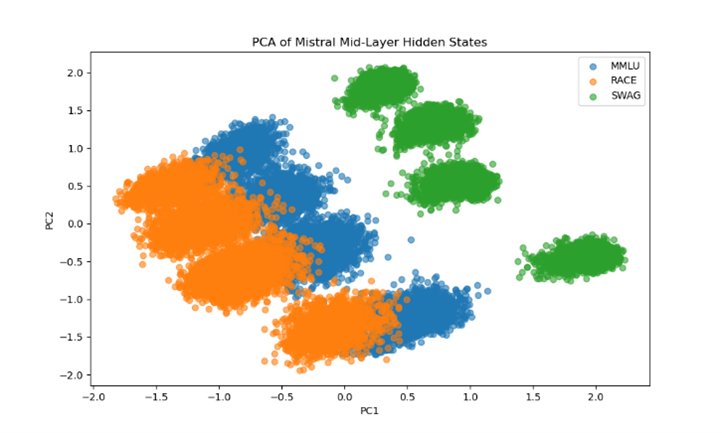}
    \caption{Multiple-choice QA}
    \label{fig:pca-mcqa}
  \end{subfigure}
  \hfill
  \begin{subfigure}[b]{0.48\textwidth}
    \centering
    \includegraphics[width=\textwidth]{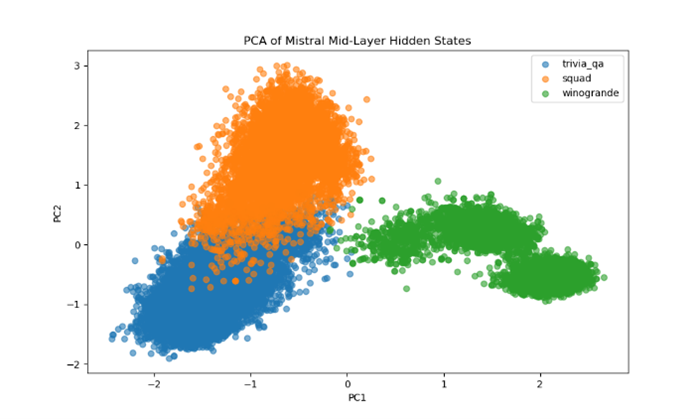}
    \caption{Short-form QA}
    \label{fig:pca-sfqa}
  \end{subfigure}

  \caption{PCA of the hidden states for multiple choice question answering (left) and short-form question answering datasets (right) using Mistral – 7B as the LLM.}
  \label{fig:pca-mistral}
\end{figure}

\section{Related Work}

\subsection{Uncertainty Quantification for LLMs Using Black-box Methods}

Black-box approaches estimate model uncertainty exclusively from the LLM’s input–output behaviour, making them suitable when internal activations are inaccessible. A prominent example is \textbf{verbalised confidence}, in which the prompt is augmented with a request for a confidence score alongside the generated answer. Although this technique is straightforward to implement, it often exhibits poor calibration: LLMs tend to assign high confidence to incorrect responses – a phenomenon known as “overconfidence” \citep{Lin2022teaching, Xiong2023confidence}. Another notable line of work that belongs to this category is \textbf{self-consistency}. This black-box method estimates confidence by sampling multiple outputs for the same prompt and measuring their agreement: high agreement implies high confidence, whereas divergent responses indicate low confidence \citep{Wang2022selfconsistency, Xiong2023confidence, Mahaut2024factual}.

\subsection{Uncertainty Quantification for LLMs Using White-box Methods}

White-box methods estimate model uncertainty by inspecting internal signals – either the probabilities that the model assigns to each generated token or the activations of its hidden layers. There are two main categories: logit-based and internal-based. The \textbf{logit-based methods} focus on assessing the sentence uncertainty using token-level probabilities or entropy \citep{Huang2023survey}. Prominent works include \cite{kadavath2022language}, in which the LLM first generates a response and then evaluates its own output as either True or False. The confidence level is determined by the probability the model assigns to the True label during this self-evaluation, \cite{farquhar2024detecting} proposed semantic uncertainty, which estimates semantic entropy by first grouping semantically equivalent samples using bidirectional entailment and then summing the predicted probabilities within each cluster, and \cite{duan2023shifting} proposed shifting attention to more relevant tokens, resolving the problem of tokens with limited semantics being equally or excessively weighted. Despite being effective, these methods primarily capture how fluently a claim is made rather than its factual reliability \citep{Mahaut2024factual}. \textbf{Internal-based methods}, by contrast, exploit the hidden representations of the model to train external estimators (probes) that predict uncertainty directly from internal activations \citep{Azaria2023internalstate, Beigi2024internalinspector, He2024factoscope, Liu2024uncertainty}. These methods have been shown to capture richer signals that are correlated with truthfulness or factual reliability \citep{Azaria2023internalstate, Orgad2024intrinsic}. However, the trained probes often overfit to the domains on which they were trained and fail to generalise to new tasks or data distributions. One recent work by \cite{kossen2024semantic} also utilises the hidden states of LLMs, but for predicting semantic entropy. The main difference lies in how the labels for training the probe are constructed. In their approach, labels are generated by sampling multiple responses from the LLM and computing the semantic entropy, rather than directly evaluating response accuracy against ground truth labels. The semantic entropy probe has been shown to outperform the standard accuracy probe trained with ground truth supervision in terms of generalisation performance. In our work, however, we use ground truth labels, as they are readily available and to avoid the resource-intensive process of sampling multiple responses for probe training. Additionally, our primary focus is on improving the generalisation performance of the accuracy probe, and we leave the comparison with the semantic entropy probe to future work. 

\subsection{Generalisation In Trained Probe}

To overcome the generalisation gap in the trained probe method, several lines of work have proposed different techniques to tackle the generalisation issue. Specifically, \cite{Liu2024hyperplane} proposed to train the probe on a diverse set of datasets covering multiple tasks and domains to find out a universal truthfulness hyperplane that can generalise across tasks and datasets. Even though the results showed decent generalisation performance, they have yet to verify whether this universal truthfulness hyperplane is causally related to model behaviour. In addition, this technique requires collecting an extensive source of different datasets, even if only a few datapoints are required from each, it is still considered to be resource intensive. As a result, \cite{Zhang2024promptguided} proposed to reframe every statement as a True/False question and trained probes on the answer token’s hidden states. Even though this method is proved to be effective in improving generalisation performance, it is only limited to True/False statements, making it difficult to be applied in real world situations. Our method sidesteps both limitations by combining hidden‐state features with data‐agnostic metrics, such as token-probability and entropy statistics, in a single probe. The reason behind this is simple, hidden-state features carry rich information related to the truthfulness of the answers generated, which has been proved by prior studies, while the data-agnostic features help improve the generalisation performance of the trained probe as they are task-independent \citep{He2024factoscope}. With this combination, we aim to examine the effectiveness of the probe in out-of-distribution cases, thereby helping to address its generalisation limitations. 

\section{Conclusion}

The internal-based method, which utilises the hidden states of LLMs to predict factual accuracy, is one of the methods that has shown strong performance and is thus a focus of this paper. However, a key limitation of this approach is its poor generalisation across tasks and datasets. To address this, we proposed incorporating data-agnostic features to examine the improvements in generalisation performance of the trained probe across different tasks and datasets. Our experiments demonstrate that integrating probability- and entropy-based metrics significantly enhances performance, particularly on short-form question answering benchmarks. We further investigate feature selection by retaining only the top-k most informative hidden-state dimensions, which does help amplify the contribution of data-agnostic signals and yield additional gains in cross-domain settings. However, the trained probe still underperforms on certain transfer pairs that involve Commonsense Reasoning datasets; further analysis reveals that in such cases, it fails to prioritise the data-agnostic features that are critical for robust, out-of-distribution generalisation.

\section{Limitation}

We acknowledge that our research still has some limitations. First, we randomly select the data-agnostic features without looking into whether all of them are effective, and whether including more or less data-agnostic features will affect the generalisation performance. Second, although the probe improved the accuracy in certain cases where data-agnostic features are included, the rate of correct answers being misclassified is relatively high – a phenomenon warranting further investigation. Third, our baseline (without data-agnostic features) drifts slightly across the three probe configurations. For example, in the MMLU-RACE case with Llama2 – 7B, baseline accuracy is 0.6095 for the one-middle-layer probe, 0.6105 for the selected-feature probe, and 0.6100 for the five-middle-layer probe. Although these differences are negligible, they introduce instability in our baseline and might make the comparisons inconsistent. Finally, the probe underweights data-agnostic features in Commonsense Reasoning datasets, suggesting that their hidden states may already encode similar information. Future work should systematically examine this phenomenon, paving the way for effective utilisation of data-agnostic features in generalisation performance improvement.

\nocite{*}
\bibliographystyle{apalike}
\bibliography{references}

\end{document}